# Dialogue System: A Brief Review

**Suket Arora[1], Kamaljeet Batra[2], Sarabjit Singh[3]**

[1]Research Scholar, Punjab Technical University, Kapoorthala (suket.arora@yahoo.com)

[2]Assistant. Professor, PG Department of CS & IT, DAV College, Amritsar (kamaljit_batra@yahoo.com)

[3]Assistant Professor, Department of Languages & Translation,
Punjab Technical University, Kapoorthala (mannrabia@gmail.com)

**Abstract**

*A Dialogue System is a system which interacts with human in natural language. At present many universities are developing the dialogue system in their regional language. This paper will discuss about dialogue system, its components, challenges and its evaluation. This paper helps the researchers for getting info regarding dialogues system.*

## 1. Introduction

A Dialogue is a conversation between two or more agents, be they human or machine. Research on dialogue is on two topics human-human dialogue and human-computer dialogue. The later is involved in a Dialogue System, a computerized system whose aim is to interact with humans in a natural language. Today dialogue system is developing in text, graphical, spoken and multimodal systems.

## 2. Dialogue System

A dialogue system is a computer program that communicates with a human user in a natural way.[1] The dialogue System provides an interface between the user and a computer-based application that permits interaction with the application in a relatively natural manner. The System can be CUI, GUI, VUI and multi model etc. it can be used in telephones, PDA systems, cars, robot systems and web browsers. A text based dialogue system is in which we chat with the system. A spoken dialogue systems is defined as a computer systems that human interact on a turn-by-turn basic and in which spoken natural language interface plays an important part in the communication.[2] A multimodal dialogue systems are those which are dialogue systems that process two or more combined user input modes - such as speech, pen, touch, manual gestures, gaze, and head and body movements - in a coordinated manner with multimedia system output.[3] Different Dialogue Systems have different architectures but they have same set of phases which are Input Recognition, Natural Language Understanding, Dialogue Management, Response Generation and Output Renderer.[4]

## 3. Components of Dialogue System

A Dialogue system has mainly seven components.[5] These components are following:

- **Input Decoder**
- **Natural Language Understanding**
- **Dialogue Manager**
- **Domain Specific Component**
- **Response Generator**
- **Output Renderer**

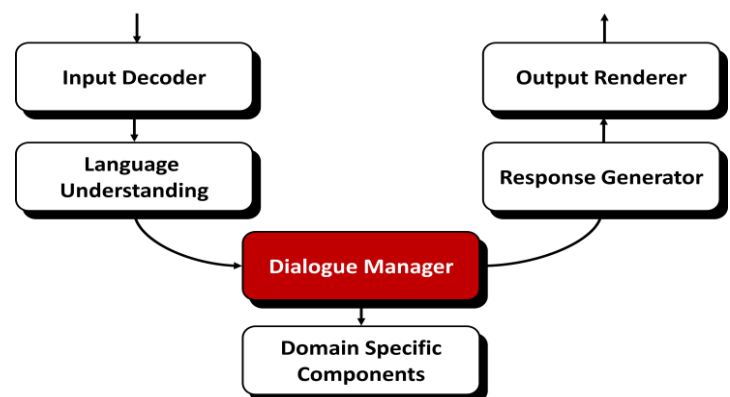

**Fig.1 - Components of dialogue system**

## 3.1 Input Decoder

Input Decoder component is the one which recognizes the input. It converts the input to the simple text. This component is present only in which are not text base dialogue systems. This component involves conversion of spoken sound (user utterances) to text (a string of words). This requires the knowledge of phonetics and phonology. Phonetics is branch of linguistic which deals with the sound of speech and their production, combination, description and representation by written symbols. Phonology is study of speech sound in language or a language with reference to their distribution and patterning and to tacit rules governing pronunciation. For this purpose speech Recognition is needed. There are many systems available for this purpose. These are called Automatic Speech Recognition (ASR), Computer Speech Recognition or simply Speech to Text (STT). Besides speech the dialogue system can have other inputs like gesture, handwriting etc.

## 3.2 Natural Language Understanding

As the name suggest this unit try to understand what user want to tell. It converts the sequence of words into a semantic representation that can be used by the dialogue manager. This component involves use of morphology, syntax and semantics. Morphology is the study of the structure and content of word forms. After identifying the keywords and forming a meaning it provide it to dialogue manager.

## 3.3 Dialogue Manager

The Dialogue Manager manages all aspects of the dialogue. It takes a semantic representation of the user's text, figures out how text fits in the overall context and creates a semantic representation of the system response. It performs many tasks these are:

- Maintains the history of dialogue
- Adopts certain dialogue strategies
- Deal with malformed and unrecognized text
- Retrieve the contents stored in files or database
- Decides the best response for user
- Manage initiative and system response
- Handle issue of pragmatics
- Discourse analysis
- It also performs grounding

For these tasks dialogue manager has many components these components are:

- Dialogue Model
- User Model
- Knowledge Base
- Discourse Manager
- Reference Resolver
- Grounding Module

## 3.4 Domain Specific Component

The Dialogue Manager usually needs to interface with some external software such as a database or an expert system. The query or plans thus have to be converted from the internal representation used by the dialogue manager to the format used by the external domain specific system (e.g. SQL). This interfacing is handled by the domain specific components. This can be handled by Natural Language Query Processing system. This system generate SQL query from natural language.

## 3.5 Response Generator

This component involves constructing the message that is to be given by the user. It takes decision regarding what information should be included, how information should be structured, choice of words and syntactic structure for message. Current systems use simple methods such as insertion of retrieved data into predefined slots in a template.

## 3.6 Speech Generation

It translates the message constructed by the response generation component into spoken form. For speech generation two approaches may be used. The first approach is to use prerecorded canned speech may be used with spaces to be filled by retrieved or previously recorded samples e.g. *"Welcome, how can I help you."* The second approach is use text to speech synthesis. In this speech is generated of text. It is called Contaminative Speech Synthesis, Text to Phoneme conversion and Phoneme to speech conversion or Text to Speech (TTS).

## 4. Classification of Dialogue System

On the basis of method use to control dialogue a dialogue system can be classified in three categories[6]:

- Finite State (or graph) based systems
- Frame based systems
- Agent based systems

### 4.1 Finite State based Systems

In these types of systems the user is taken through a dialogue consisting of a sequence of predetermined steps or stages. The flow of dialogue is specified as a set of dialogue states. Following is the example:

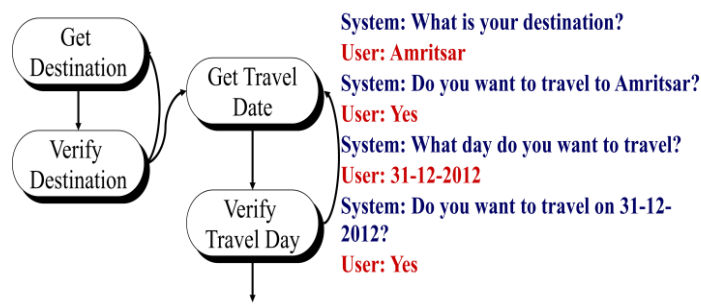

Fig. 2 – Example of Finite State based System

#### 4.1.1 Advantages

- Simple to construct
- The required vocabulary and grammar for each state can be determined in advance.

#### 4.1.2 Disadvantages

- Dialogues are not natural
- Do not allow over-informative answers
- Inhabits the user ability to ask questions and take initiative.

### 4.2 Frame Based Systems

Frame Based systems uses template filling from user response. In this system user is asked questions that enable the system to fill slots in a template in order to perform tasks. The flow of dialogue is not predetermined but depends upon the content of user input and the information the user has to elicit. Example is shown in Fig.3, in which we can see there are two different dialogues in first the dialogue goes like finite state system. In second dialogue user provide over information in respond to a question but system fills its slots of from the user's input and asks for the remaining information. This is how frame based systems works.

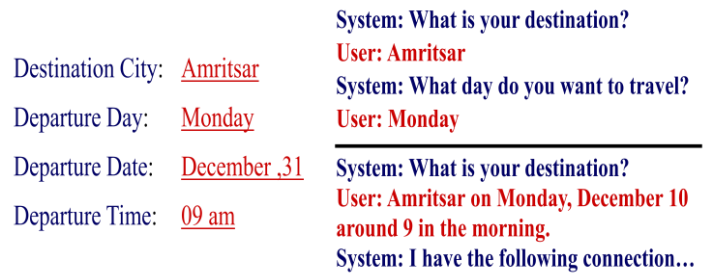

Fig. 3 – Example of Frame Based System

#### 4.2.1 Advantages

- Allow more natural Dialogues
- User can provide over informative answers

#### 4.2.2 Disadvantages

- These systems can't handle complex dialogues
- Range of application is limited to the systems that elicit information from users and act on the basis on the same

### 4.3 Agent Based Systems

These systems allow complex communication between the system, the user and the application in order to solve some problem or task. The interaction is viewed as interaction between two agents, each of which is capable of reasoning about its own actions and beliefs. The dialogue model takes the preceding context into account. The dialogue evolves dynamically as a sequence of related steps that build on top of each other.

#### 4.3.1 Advantages

- Allow natural language in complex domain
- User friendly, like talking to human

### 4.3.2 Disadvantages
- These systems are hard to build
- The agent itself are usually very complex

## 5. Challenges

To build dialogue system developers faces many difficulties. These are due lack of computer's understanding of natural language. This problem arises many challenges for developers e.g. problem of Anaphora Resolution, Inferences, Ellipsis, Pragmatics, Reference resolutions and Clarifications, Inter sentential Ellipsis etc.[7] Besides these language problem other challenges is to design system prompts, grounding, detection of conflicts and plan recognition etc. In the spoken dialogue systems the problem related to utterance of the user occur like ill formed utterances. These are the some of the challenges that developers have to take care of at designing time.

## 6. Dialogue System Evaluation

An optimal dialogue system is one which allows a user to accomplish their goals with the fewest problems. It includes evaluating the correctness of the total solution e.g. percentage of slots that were filled with the correct values. Then measures of the system's efficiency at helping users e.g. check the total elapsed time for the dialogue. In the last user's perception of the system e.g. number of times the ASR system failed to return any sentence or rejected prompts ("I'm sorry I didn't understand that"). Or the number of times the user had to barge-in (interrupt the system). We can also do user satisfaction survey. In the survey we can ask performance of ASR, TTS from user. We can take ratings for task ease, system response, interaction speed, user expertise and system behavior etc. For this survey following are some questions that can be used.

| | |
|---|---|
| TTS Performance | Was the system easy to understand? |
| ASR Performance | Did the system understand what you said? |
| Task Ease | Was it easy to find the message/flight/train you wanted? |
| Interaction Pace | Was the pace of interaction with the system appropriate? |
| User Expertise | Did you know what you could say at each point? |
| System Response | How often was the system sluggish and slow to reply to you? |
| Expected Behavior | Did the system work the way you expected it to? |
| Future Use | Do you think you'd use the system in the future? |

Fig. 4 – User Satisfaction Survey[7]

## 7. Conclusion

In the conclusion we can say that a Dialogue system is very good tool for interaction between any application and user. This is good tool and can be used in many devices such as mobiles, telephones, computers etc. It can be a good tool for web site assistance which can have domain of online shopping, travelling information, counseling, tutoring system, ticket booking, Remote banking, Travel reservation, Information enquiry, Stock transactions, Taxi bookings, Route planning etc.